\documentclass[runningheads]{llncs}
\usepackage{graphicx}

\usepackage{tikz}
\usepackage{comment}
\usepackage{amsmath,amssymb} 
\usepackage{color}

\usepackage[accsupp]{axessibility}  

\usepackage[pagebackref,breaklinks,colorlinks,bookmarks=false]{hyperref}

\usepackage[capitalize]{cleveref}
\crefname{section}{Sec.}{Secs.}
\Crefname{section}{Section}{Sections}
\Crefname{table}{Table}{Tables}
\crefname{table}{Table}{Tables}

\usepackage{multirow}
\usepackage{makecell}
\usepackage{microtype}
\usepackage{userstyle}

\begin{document}
\pagestyle{headings}
\mainmatter
\def\ECCVSubNumber{3132}  

\title{Learning Pedestrian Group Representations for Multi-modal Trajectory Prediction} 

\titlerunning{GP-Graph: Learning Group Representations for Trajectory Prediction}
%
\author{Inhwan Bae \and
Jin-Hwi Park \and
Hae-Gon Jeon\thanks{Corresponding author}}
\authorrunning{I. Bae et al.}
%
\institute{AI Graduate School, GIST, South Korea\\
\{inhwanbae, jinhwipark\}@gm.gist.ac.kr, haegonj@gist.ac.kr}
\maketitle

\begin{abstract}
Modeling the dynamics of people walking is a problem of long-standing interest in computer vision. Many previous works involving pedestrian trajectory prediction define a particular set of individual actions to implicitly model group actions. In this paper, we present a novel architecture named GP-Graph which has collective group representations for effective pedestrian trajectory prediction in crowded environments, and is compatible with all types of existing approaches. A key idea of GP-Graph is to model both individual-wise and group-wise relations as graph representations. To do this, GP-Graph first learns to assign each pedestrian into the most likely behavior group. Using this assignment information, GP-Graph then forms both intra- and inter-group interactions as graphs, accounting for human-human relations within a group and group-group relations, respectively. To be specific, for the intra-group interaction, we mask pedestrian graph edges out of an associated group. We also propose group pooling\&unpooling operations to represent a group with multiple pedestrians as one graph node. Lastly, GP-Graph infers a probability map for socially-acceptable future trajectories from the integrated features of both group interactions. Moreover, we introduce a group-level latent vector sampling to ensure collective inferences over a set of possible future trajectories. Extensive experiments are conducted to validate the effectiveness of our architecture, which demonstrates consistent performance improvements with publicly available benchmarks.
Code is publicly available at \url{https://github.com/inhwanbae/GPGraph}.

\end{abstract}

\section{Introduction}
Pedestrian trajectory prediction attempts to forecast the socially-acceptable future paths of people based on their past movement patterns. These behavior patterns often depend on each pedestrian's surrounding environments, as well as collaborative movement, mimicking a group leader, or collision avoidance. Collaborative movement, one of the most frequent patterns, occurs when several colleagues form a group and move together. 
Computational social scientists estimate that up to 70\% of the people in a crowd will form groups~\cite{moussaid2010walking,rudenko2018human}. They also gather surrounding information and have the same destination~\cite{moussaid2010walking}. Such groups have characteristics that are distinguishable from those of individuals, maintain rather stable formations, and even provide important cues that can be used for future trajectory prediction~\cite{rudenko2018human,zhou2012understanding}.

Pioneering works in human trajectory forecasting model the group movement by assigning additional hand-crafted terms as energy potentials~\cite{Pellegrini,yamaguchi2011you,robicquet2016learning}. These works account for the presence of other group members and physics-based attractive forces, which are only valid between the same group members. In recent works, convolutional neural networks (CNNs) and graph neural networks (GNNs) show impressive progress modeling the social interactions, including traveling together and collision avoidance~\cite{alahi2016social,gupta2018social,mohamed2020social,Bae_Jeon_2021,Shi2021sgcn}. Nevertheless, trajectory prediction is still a challenging problem because of the complexity of implicitly learning individual and group behavior at once.

\begin{figure}[t]
\centering
\includegraphics[width=\columnwidth]{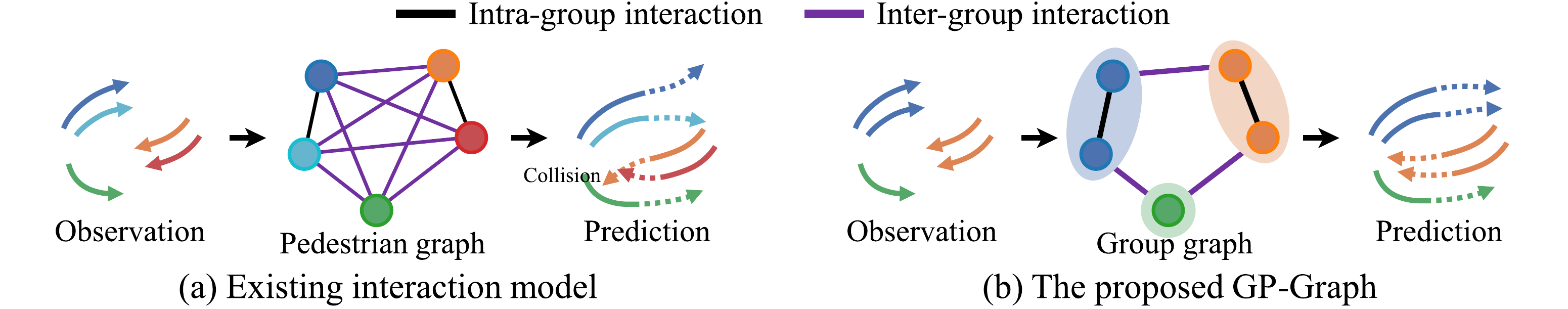}
\vspace{-7mm}
\caption{Comparison of existing agent-agent interaction graphs and the proposed group-aware GP-Graph. To capture social interactions, (a) existing pedestrian trajectory prediction models each pedestrian on a graph node. Since the pedestrian graph is a complete graph, it is difficult to capture the group's movement because it becomes overly complex in a crowded scene. (b) GP-Graph is directly able to learn an intra-/inter-group interaction while keeping the agent-wise structure.}
\vspace{-2mm}
\label{fig:figure1}
\end{figure}

There are several attempts that explicitly encode the group coherence behaviors by assigning hidden states of LSTM with a summation of other agents' states, multiplied by a binary group indicator function~\cite{bisagno2018group}. However, existing studies have a critical problem when it comes to capturing the group interaction. Since their forecasting models focus more on individuals, the group features are shared at the individual node as illustrated in~\cref{fig:figure1}(a). Although this approach can conceptually capture group movement behavior, it is difficult for the learning-based methods to represent it because of the overwhelming number of edges for the individual interactions. And, this problem is increasingly difficult in crowded environments.

To address this issue, we propose a novel general architecture for pedestrian trajectory prediction: GrouP-Graph (GP-Graph). As illustrated in~\cref{fig:figure1}(b), our GP-Graph captures intra-(members in a group) and inter-group interactions by disentangling input pedestrian graphs. Specifically, our GP-Graph first learns to assign each pedestrian into the most likely behavior group. The group indices of each pedestrian are generated using a pairwise distance matrix. To make the indexing process end-to-end trainable, we introduce a straight-through group back-propagation trick inspired by the Straight-Through estimator~\cite{bengio2013estimating,jang2016categorical,maddison2016concrete}. 
Using the group information, GP-graph then transforms the input pedestrian graph into both intra- and inter-group interaction graphs. We construct the intra-group graph by masking out edges of the input pedestrian graph for unassociated group members. For the inter-group graph, we propose group pooling\&unpooling operations to represent a group with multiple members as one graph node. 
By applying these processes, GP-Graph architecture has three advantages: (1) It reduces the complexity of trajectory prediction which is caused by the different social behaviors of individuals, by modeling group interactions. (2) It alleviates inherent scene bias by considering the huge number of unseen pedestrian graph nodes between the training and test environments, as discussed in~\cite{liu2021causal}. (3) It offers a graph augmentation effect with pedestrian node grouping.

Next, through weight sharing with baseline trajectory predictors, we force a hierarchy representation from both the input pedestrian graph and the disentangled interactions. This representation is used to infer a probability map for socially-acceptable future trajectories after passing through our group integration module. In addition, we introduce a group-level latent vector sampling to ensure collective inferences over a set of plausible future trajectories.

To the best of our knowledge, this is the first model that literally pools pedestrian colleagues into one group node to efficiently capture group motion behaviors, and learns pedestrian grouping in an end-to-end manner. Furthermore, GP-Graph has the best performance on various datasets among existing methods when unifying with GNN-based models, and it can be integrated with all types of trajectory prediction models, achieving consistent improvements. We also provide extensive ablation studies to analyze and evaluate our GP-Graph.

\vspace{-1.5mm}
\section{Related Works}
\vspace{-0.5mm}
\subsection{Trajectory Prediction}
Earlier works~\cite{helbing1995social,5459260,5206641,yamaguchi2011you} model human motions in crowds using hand-crafted functions to describe attractive and repulsive forces. Since then, pedestrian trajectory prediction has been advanced by research interest in computer vision. Such research leverages the impressive capacity of CNNs which can capture social interactions between surrounding pedestrians. One pioneering work is Social-LSTM~\cite{alahi2016social}, which introduces a social pooling mechanism considering a neighbor's hidden state information inside a spatial grid. Much of the emphasis in subsequent research has been to add human-environment interactions from a surveillance view perspective~\cite{sadeghian2019sophie,liang2019peeking,kosaraju2019social,sun2020reciprocal,dendorfer2021mggan,zhao2019matf,tao2020dynamic,sun2020rsbg,Marchetti_2020_CVPR,shafiee2021Introvert}. Instead of taking environmental information into account, some methods directly share hidden states of agents between other interactive agents~\cite{gupta2018social,vemula2018social,salzmann2020trajectron++}. In particular, Social-GAN~\cite{gupta2018social} takes the interactions via max-pooling in all neighborhood features in the scene, and Social-Attention~\cite{vemula2018social} introduces an attention mechanism to impose a relative importance on neighbors and performs a weighted aggregation for the features. 

In terms of graph notations, each pedestrian and their social relations can be represented as a node and an edge, respectively.
When predicting pedestrian trajectories, graph representation is used to model social interactions with graph convolutional networks (GCNs)~\cite{kipf2016semi,mohamed2020social,sun2020rsbg,Bae_Jeon_2021}, graph attention networks (GATs) \cite{velivckovic2018graph,huang2019stgat,kosaraju2019social,liang2020garden,Shi2021sgcn,bae2022npsn}, and transformers~\cite{yu2020spatio,yuan2021agent,gu2022mid}. Usually, these approaches infer future paths through recurrent estimations~\cite{alahi2016social,gupta2018social,salzmann2020trajectron++,zhao2021experttraj,chen2021disdis,lee2022musevae,gu2022mid} or extrapolations~\cite{mohamed2020social,Bae_Jeon_2021,Shi2021sgcn,li2021stcnet}. 
Other types of relevant research are based on probabilistic inferences for multi-modal trajectory prediction using Gaussian modeling~\cite{alahi2016social,Bae_Jeon_2021,mohamed2020social,shi2020multimodal,yu2020spatio,li2020Evolvegraph,Shi2021sgcn,xu2022tgnn}, generative models \cite{gupta2018social,sadeghian2019sophie,kosaraju2019social,zhao2019matf,sun2020reciprocal,dendorfer2021mggan,huang2019stgat}, and a conditional variational autoencoder~\cite{Lee_2017_CVPR,li2019conditional,Ivanovic_2019_ICCV,salzmann2020trajectron++,mangalam2020pecnet,chen2021disdis,sun2021pccsnet,lee2022musevae}. We note that these approaches focus only on learning implicit representations for group behaviors from agent-agent interactions.

\vspace{-1.5mm}
\subsection{Group-aware Representation}
\vspace{0.5mm}
Contextual and spatial information can be derived from group-aware representations of agent dynamics. To accomplish this, one of the group-aware approaches is social grouping, which describes agents in groups that move differently than independent agents.

In early approaches~\cite{Group_Kmeans,Group_SVC,coherentfilter}, pedestrians can be divided into several groups based on behavior patterns. To represent the collective activities of agents in a supervised manner, a work in \cite{Pellegrini} exploits conditional random fields (CRF) to jointly predict the future trajectories of pedestrians and their group membership. 
Yamaguchi~\etal~\cite{yamaguchi2011you} harness distance, speed, and overlap time to train a linear SVM to classify whether two pedestrians are in the same group or not. 
In contrast, a work in \cite{Ge_hierarchical_Group} proposes automatic detection for small groups of individuals using a bottom-up hierarchical clustering with speed and proximity features.

Group-aware predictors recognize the affiliations and relations of individual agents, and encode their proper reactions to moving groups. Several physics-based techniques represent group relations by adding attractive forces among group members~\cite{yamaguchi2011you,Pellegrini,Robicquet,moussaid2010walking,Qiu_Hu2010,Seitz2012,Singh2009}.
Although a dominant learning paradigm \cite{alahi2016social,zhang2019,Pfeiffer2018,Varshneya2017,Bartoli2018} implicitly learns intra- and inter-group coherency, only two works in~\cite{bisagno2018group,Fernando2018} explicitly define group information.
To be specific, one~\cite{bisagno2018group} identifies pedestrians walking together in the crowd using a coherent filtering algorithm~\cite{coherentfilter}, and utilizes the group information in a social pooling layer to share their hidden states. Another work~\cite{Fernando2018} proposes a generative adversarial model (GAN)-based trajectory model, jointly learning informative latent features for simultaneous pedestrian trajectory forecasting and group detection. These approaches only learn individual-level interactions within a group, but do not encode their affiliated groups and future paths at the same time. Unlike them, our GP-Graph aggregates a group-group relation via a novel group pooling in the proposed end-to-end trainable architecture without any supervision.

\vspace{-1.5mm}
\subsection{Graph Node Pooling}
\vspace{0.5mm}
Pooling operations are used for features extracted from grid data, like images, as well as graph-structured data. 
However, there is no geographic proximity or order information in the graph nodes that existing pooling operations require. 
As alternative methods, three types of graph pooling are introduced: topology-based pooling~\cite{topology_defferrard2016,topology_Rhee2018}, global pooling~\cite{global_Gilmer2017,global_Zhang2018}, and hierarchical pooling~\cite{hierarchcal_Cangea2018,hierarchcal_Gao2019,hierarchcal_Ying2018}. These approaches are designed for general graph structures. However, since human behavior prediction has time-variant and generative properties, it is no possible to leverage the advantages of these pooling operations for this task.

\begin{figure}[t]
\centering
\vspace{1mm}
\includegraphics[width=\columnwidth]{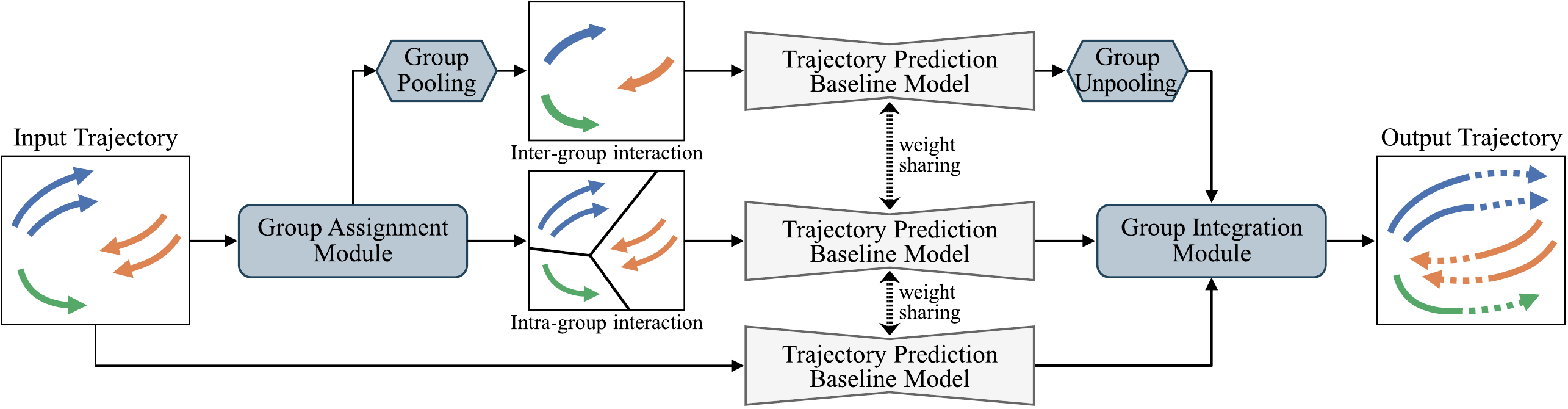}
\vspace{-7.5mm}
\caption{An overview of our GP-Graph architecture. Starting with graph-structured trajectories for $N$ pedestrians, we first estimate grouping information with the Group Assignment Module. We then generate both intra-/inter-group interaction graphs by masking out unrelated nodes and by performing pedestrian group pooling. The weight-shared trajectory prediction model takes the three types of graphs and capture group-aware social interactions. Group pooling operators are then applied to encode agent-wise features from group-wise features, and then fed into the Group Integration Module to estimate the probability distribution for future trajectory prediction.}
\vspace{-1mm}
\label{fig:figure3}
\end{figure}

\section{Proposed Method}
\vspace{-1mm}
In this work, we focus on how group awareness in crowds is formed for pedestrian trajectory prediction. We start with a definition of a pedestrian graph and trajectory prediction in~\cref{sec:problem_definition}. We then introduce our end-to-end learnable pedestrian group assignment technique in~\cref{sec:group_prediction}. Using group index information and our novel pedestrian group pooling\&unpooling operations, we construct a group hierarchy representation of pedestrian graphs in~\cref{sec:trajectory_prediction}. The overall architecture of our GP-Graph is illustrated in~\cref{fig:figure3}.

\vspace{-2mm}
\subsection{Problem Definition}
\vspace{-0mm}
\label{sec:problem_definition}
Pedestrian trajectory prediction can be defined as a sequential inference task made observations for all agents in a scene. Suppose that $N$ is the number of pedestrians in a scene, the history trajectory of each pedestrian $n \in [1, ..., N]$ can be represented as $\bm{X}_n\!=\!\{ (x_n^t, y_n^t)\,|\,t\!\in\![1, ..., T_{obs}] \}$, where the $(x_n^t, y_n^t)$ is the 2D spatial coordinate of a pedestrian $n$ at specific time $t$. Similarly, the ground truth future trajectory of pedestrian $n$ can be defined as $\bm{Y}_n\!=\!\{ (x_n^t, y_n^t)\,|\,t\!\in\![T_{obs}\!+\!1, ..., T_{pred}] \}$. 

The social interactions are modeled from the past trajectories of other pedestrians. In general, the pedestrian graph $\mathcal{G}_{ped}\!=\!(\mathcal{V}_{ped}, \mathcal{E}_{ped})$ refers to a set of pedestrian nodes $\mathcal{V}_{ped} = \{ \bm{X}_n\,|\,n\!\in\![1, ..., N] \}$ and edges on their pairwise social interaction $\mathcal{E}_{ped} = \{ e_{i,j}\,|\,i,j\!\in\![1, ..., N] \}$. The trajectory prediction process forecasts their future sequences based on their past trajectory and the social interaction as:
\begin{equation}
    \widehat{\bm{Y}} = F_\theta \left(X,\,\mathcal{G}_{ped}\right)
\end{equation}
where $\widehat{\bm{Y}} = \{ \widehat{\bm{Y}}_n\,|\,n\!\in\![1, ..., N] \}$ denotes the estimated future trajectories of all pedestrians in a scene, and $F_\theta(\,\cdot\,)$ is the trajectory generation network.

\vspace{-1mm}
\subsection{Learning the Trajectory Grouping Network}
\vspace{-0mm}
\label{sec:group_prediction}
Our goal in this work is to encode powerful group-wise features beyond existing agent-wise social interaction aggregation models to achieve highly accurate human trajectory prediction. The group-wise features represent group members in input scenes as single nodes, making pedestrian graphs simpler. We use a U-Net architecture with pooling layers to encode the features on graphs. By reducing the number of nodes through the pooling layers in the U-Net, higher-level group-wise features can be obtained. After that, agent-wise features are recovered through unpooling operations.

Unlike conventional pooling\&unpooling operators working on grid-structured data, like images, it is not feasible to apply them to graph-structured data. Some earlier works to handle this issue~\cite{hierarchcal_Cangea2018,hierarchcal_Gao2019}. 
The works focus on capturing global information by removing relatively redundant nodes using a graph pooling, and restoring the original shapes by adding dummy nodes from a graph unpooling if needed.
However, in pedestrian trajectory prediction, each node must keep its identity index information and describe the dynamic property of the group behavior in scenes. 
For that, we present pedestrian graph-oriented group pooling\&unpooling methods.
We note that it is the first work to exploit the pedestrian index itself as a group representation.

\noindent\textbf{Learning pedestrian grouping.}\quad
First of all, we estimate grouping information to which the pedestrian belongs using a Group Assignment Module. Using the history trajectory of each pedestrian, we measure the feature similarity among all pedestrian pairs based on their $L_2$ distance. With this pairwise distance, we pick out all pairs of pedestrians that are likely to be a colleague (affiliated with same group). The pairwise distance matrix $\bm{D}$ and a set of colleagues indices $\Upsilon$ are defined as:
\begin{equation}
    \bm{D}_{\,i,j} = \| F_\phi (\bm{X}_i) - F_\phi (\bm{X}_j)\| ~~~\text{for}~~ i,j \in [1, ..., N],
    \label{eq:pair_distance}
    \vspace{-3mm}
\end{equation}
\begin{equation}
    \Upsilon = \{ \text{pair}(i,\,j)\,|\,i,j \in [1, ..., N], ~i \neq j, ~\bm{D}_{\,i,j} \leq \pi \},
    \vspace{3mm}
\end{equation}
where $F_\phi(\,\cdot\,)$ is a learnable convolutional layer and $\pi$ is a learnable thresholding parameter.

Next, using the pairwise colleague set $\Upsilon$, we arrange the colleague members in associated groups and assign their group index. We make a group index set $G$, which is formulated as follows:

\begin{equation}
    G = \Big\{ G_k \,|\, G_k = \!\!\bigcup_{(i,j) \in \Upsilon}\! \{i,\,j\},~~G_a\!\cap G_b = \varnothing~~\text{for}~ a \neq b \Big\}
    \label{eq:group_assign}
\end{equation}
where $G_k$ denotes the $k$-th group and is the union of each pair set $(i,j)$. This information is used as important prior knowledge in the subsequent pedestrian group pooling and unpooling operators.

\noindent\textbf{Pedestrian group pooling.}\quad
Based on the group behavior property that group members gather surrounding information and share behavioral patterns, we group the pedestrian nodes, where the corresponding node's features are aggregated into one node. The aggregated group features are then stacked for subsequent social interaction capturing modules (\ie GNNs). Here, the most representative feature for each pedestrian node is selected via an average pooling.  With the feature, we can model the group-wise graph structures, which have much fewer number of nodes than the input pedestrian graph, as will be demonstrated in~\cref{sec:qualitative}. 
We define the pooled group-wise trajectory feature $\bm{Z}$ as follows:
\begin{equation}
    \bm{Z} = \{\bm{Z}_k\,|\,k \in [1, ..., K]\}, ~~~~~\bm{Z}_k = \frac{1}{|G_k|} \sum_{i\;\!\in\;\!G_k} \!\bm{X}_i,
\end{equation}
where $K$ is the total group numbers in $G$.

\noindent\textbf{Pedestrian group unpooling.}\quad
Next, we upscale the group-wise graph structures back to their original size by using an unpooling operation. This enables each pedestrian trajectory to be forecast with output agent-wise feature fusion information. In existing methods~\cite{hierarchcal_Cangea2018,hierarchcal_Gao2019}, zero vector nodes are appended into the group features during unpooling. 
The output of the convolution process on the zero vector nodes fails to exhibit the group properties. To alleviate this issue, we duplicate the group features and then assign them into nodes for all the relevant group members so that they have identical group behavior information. The pedestrian group unpooling operator can be formulated as follows:
\begin{equation}
    \widebar{\bm{X}} = \{ \widebar{\bm{X}}_n\,|\,n \in [1, ..., N] \}, ~~~~~\widebar{\bm{X}}_n = \bm{Z}_k ~~~\text{where}~~ n \in G_k,
\end{equation}
where $\widebar{\bm{X}}$ is the agent-wise trajectory feature reconstructed from $Z$, having the same order of pedestrian indices as in $\bm{X}$.

\noindent\textbf{Straight-Through Group Estimator.}\quad
A major hurdle, when training the group assignment module in \cref{eq:group_assign} which is a sampling function, is that index information is not treated as learnable parameters. Accordingly, the group index cannot be trained using standard backpropagation algorithms. The reason is why the existing methods utilize separate training steps  from main trajectory prediction networks for the group detection task.

We tackle this problem by introducing a Straight-through (ST) trick, inspired by the biased path derivative estimators in~\cite{bengio2013estimating,jang2016categorical,maddison2016concrete}. Instead of making the discrete index set $G_k$ differentiable, we separate the forward pass and backward pass of the group assignment module in the training process. Our intuition for constructing the backward pass is that group members have similar features with closer pairwise distance between colleagues. 

In the forward pass, we perform our group pooling over both pedestrian features and the group index from the input trajectory and estimated group assignment information, respectively. For the backward pass, we propose group-wise continuous relaxed features to approximate the group indexing process. 
We compute the probability that a pair of pedestrians belongs to the same group using the proposed differentiable binary thresholding function 
$\frac{1}{1+\exp(x-\pi)}$, and apply it on the pairwise distance matrix $\bm{D}$.
We then measure the normalized probability $\bm{A}$ of the summation of all neighbors' probability. Lastly, we compute a new pedestrian trajectory feature $\bm{X}'$ by aggregating features between group members through the matrix multiplication of $\bm{X}$ and $\bm{A}$ as follows:
\begin{equation}
    \bm{A}_{\,i,j} = \frac{\frac{1}{1 + \exp\!\big(\frac{\bm{D}_{\,i,j}-\pi}{\tau}\big)}}{\sum_{i=1}^{N} \Big(\raisebox{0.6ex}{$\frac{1}{1 + \exp\!\big(\frac{\bm{D}_{\,i,j}-\pi}{\tau}\big)}$}\Big)} ~~~\text{for}~~ i,j \in [1, ..., N],
    \vspace{-3mm}
\end{equation}
\begin{equation}
    \bm{X}' = \langle\, \bm{X} - \bm{X}\bm{A} \,\,\rangle + \bm{X}\bm{A},
    \label{eq:groupwise_cont_approx}
    \vspace{1mm}
\end{equation}
where $\tau$ is the temperature of the sigmoid function and $\langle\,\cdot\,\rangle$ is the \textit{detach} (in PyTorch) or \textit{stop gradient} (in Tensorflow) function which prevents the backpropagation. 

For further explanation of \cref{eq:groupwise_cont_approx}, we replace the input of pedestrian group pooling module $X$ with a new pedestrian trajectory feature $\bm{X}'$ in implementation. To be specific, we can remove $\bm{X}\bm{A}$ in the forward pass, allowing us to compute a loss for the trajectory feature $\bm{X}$. In contrast, due to the stop gradient $\langle\,\cdot\,\rangle$, the loss is only backpropagated to $\bm{X}\bm{A}$ in the backward pass. To this end, we can train both the convolutional layer $F_\phi$ and the learnable threshold parameter $\pi$ which are used for the computation of the pairwise distance matrix $\bm{D}$ and the construction of group index set $G$, respectively.

\begin{figure}[t]
\centering
\includegraphics[width=\columnwidth]{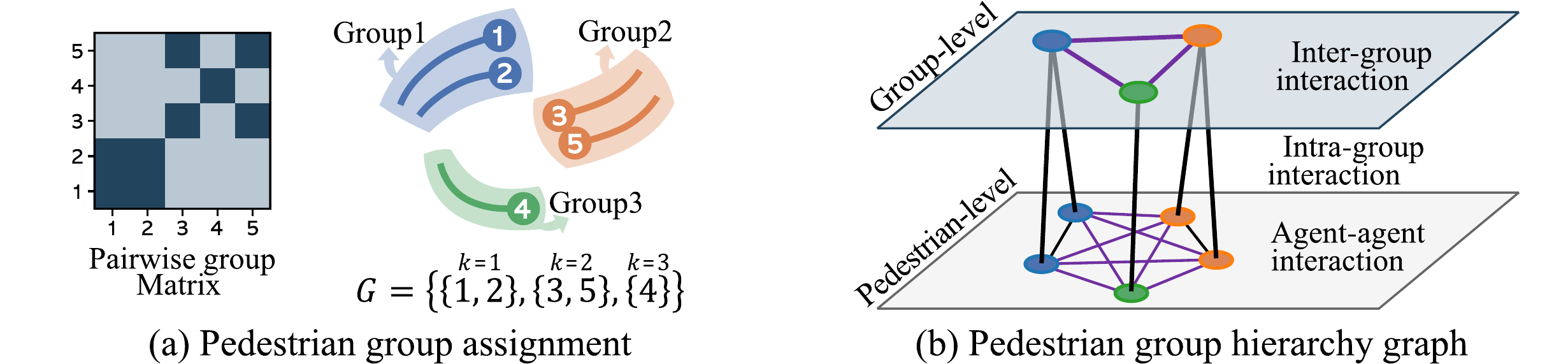}
\vspace{-7mm}
\caption{An illustration of our pedestrian group assignment method using a pairwise group probability matrix $A$. With a group index set $G$, a pedestrian group hierarchy is constructed based on three types of interaction graphs.}
\vspace{-3mm}
\label{fig:figure2}
\end{figure}

\vspace{-3mm}
\subsection{Pedestrian Group Hierarchy Architecture}
\label{sec:trajectory_prediction}
Using the estimated pedestrian grouping information, we reconstruct the initial social interaction graph $\mathcal{G}_{ped}$ in an efficient form for pedestrian trajectory prediction. Instead of the existing complex and complete pedestrian graph, intra- and inter-group interaction graphs capture the group-ware social relation, as illustrated in~\cref{fig:figure2}.

\noindent\textbf{Intra-group interaction graph.}\quad
We design a pedestrian interaction graph that captures relations between members affiliated with the same group. The intra-group interaction graph $\mathcal{G}_{member}\!=\!(\mathcal{V}_{ped}, \mathcal{E}_{member})$ consists of a set of pedestrian nodes $\mathcal{V}_{ped}$ and edges on their pairwise social interaction of group members $\mathcal{E}_{member} = \{ e_{i,j}\,|\,i,j\!\in\![1, ..., N], k\!\in\![1, ..., K], \{i,j\}\!\subset\!G_k \}$. Through this graph representation, pedestrian nodes can learn social norms of internal collision avoidance between group members while maintaining their own formations and on-going directions.

\noindent\textbf{Inter-group interaction graph.}\quad
Inter-group interactions (group-group relation) are indispensable to learn social norms between groups as well. To take various group behaviors such as following a leading group, avoiding collisions and joining a new group, we create an inter-group interaction graph $\mathcal{G}_{group}\!=\!(\mathcal{V}_{group}, \mathcal{E}_{group})$. Here, nodes refer to each group's features $\mathcal{V}_{group} = \{ \widebar{\bm{X}}_k\,|\,k\!\in\![1, ..., K] \}$ generated with our pedestrian group pooling operation, and edges mean the pairwise group-group interactions $\mathcal{E}_{group} = \{ \bar{e}_{p,q}\,|\,p,q\!\in\![1, ..., K] \}$.

\noindent\textbf{Group integration network.}\quad
We incorporate the social interactions as a form of group hierarchy into well-designed existing trajectory prediction baseline models in~\cref{fig:figure2}(b). Meaningful features can be extracted by feeding a different type of graph-structured data into the same baseline model. Here, the baseline models share their weights to reduce the amount of parameters while enriching the augmentation effect. Afterward, the output features from the baseline models are aggregated agent-wise, and are then used to predict the probability map of future trajectories using our group integration module. The generated output trajectory $\widehat{Y}$ with the group integration network $F_\psi$ is formulated as:
\begin{equation}
    \widehat{Y} = F_\psi \big( \!\!
    \underbrace{F_\theta (X\!,\mathcal{G}_{ped})}_{\substack{\text{Agent-wise}\text{ GNN}}}\!\!,~
    \underbrace{F_\theta (X\!,\mathcal{G}_{member})}_{\substack{\text{Intra-group}\text{ GNN}}}\,,~
    \underbrace{F_\theta (\widebar{X}\!,\mathcal{G}_{group})}_{\substack{\text{Inter-group}\text{ GNN}}} \big).
\end{equation}

\noindent\textbf{Group-level latent vector sampling.}\quad
\label{sec:grouplevel_sampling}
To infer the multi-modal future paths of pedestrians, an additional random latent vector is introduced with an input observation path. This latent vector becomes a factor, determining a person's choice of behavior patterns, such as acceleration/deceleration and turning to right/left. There are two ways to adopt this latent vector in trajectory generation: (1) Scene-level sampling~\cite{gupta2018social} where everyone in the scene shares one latent vector, unifying the behavior patterns of all pedestrians in a scene (\eg, all pedestrians are slow down); (2) Pedestrian-level sampling~\cite{salzmann2020trajectron++} that allocates the different latent vectors for each pedestrian, but forces the pedestrians to have different patterns, where the group behavior property is lost. 

We propose a group-level latent vector sampling method as a compromise of the two ways. We use the group information estimated from the GP-Graph to share the latent vector between groups. If two people are not associated with the same group, an independent random noise is assigned as a latent vector. In this way, it is possible to sample a multi-modal trajectory, which is independent of other groups members and follows associated group behaviors. The effectiveness of the group-level sampling is visualized in~\cref{sec:qualitative}.

\vspace{-3mm}
\subsection{Implementation Details}
To validate the generality of our GP-Graph, we incorporate it into four state-of-the-art baselines: three different GNN-based baseline methods including STGCNN (GCN-based)~\cite{mohamed2020social}, SGCN (GAT-based)~\cite{Shi2021sgcn} and STAR (Transformer-based)~\cite{yu2020spatio}, and one non-GNN model, PECNet~\cite{mangalam2020pecnet}. We simply replace their trajectory prediction parts with ours. We additionally embed our agent/intra-/inter-graphs on the baseline networks, and compute integrated output trajectories to obtain the group-aware prediction.

For our proposed modules, we initialize the learnable parameter $\pi$ as one, which cut the total number of nodes moderately down by half, with the group pooling in the initial training step. Other learnable parameters such as $F_\theta$, $F_\phi$ and $F_\psi$ are randomly initialized. We set the hyperparameter $\tau$ to 0.1 to give the binary thresholding function a steep slope.

To train the GP-Graph architecture, we use the same training hyperparameters (\eg, batch size, train epochs, learning rate, learning rate decay), loss functions, and optimizers of the baseline models. We note that we do not use additional group labels for an apple-to-apple comparison with the baseline models. Our group assignment module is trained to estimate effective groups for trajectory prediction in an unsupervised manner. Thanks to our powerful Straight-Through Group Estimator, it accomplish promising results over other supervised group detection networks~\cite{hierarchcal_Cangea2018} that require additional group labels.

\vspace{-2mm}
\section{Experiments}
\vspace{-1mm}
In this section, we conduct comprehensive experiments to verify how the grouping strategy contributes to pedestrian trajectory prediction. We first briefly describe our experimental setup (\cref{sec:experiment_setup}). We then provide comparison results with various baseline models for both group detection and trajectory prediction (\cref{sec:qualitative} and \cref{sec:quantitative}). We lastly conduct an extensive ablation study to demonstrate the effect of each component of our method (\cref{sec:ablation}).

\vspace{-3mm}
\subsection{Experimental Setup}
\label{sec:experiment_setup}
\vspace{-1mm}
\noindent\textbf{Datasets.}\quad
We evaluate the effectiveness of our GP-Graph by incorporating it into several baseline models and check the performance improvement on public datasets: ETH~\cite{5459260}, UCY~\cite{crowdsbyexample}, Stanford Drone Dataset (SDD)~\cite{robicquet2016learning}, and the Grand Central Station (GCS)~\cite{yi2015understanding} datasets.
The ETH \& UCY datasets contain five unique scenes (ETH, Hotel, Univ, Zara1 and Zara2) with 1,536 pedestrians, and the official leave-one-out strategy is used to train and to validate the models. SDD consists of various types of objects with a birds-eye view, and GCS shows highly congested pedestrian walking scenes. We use the standard training and evaluation protocol~\cite{gupta2018social,huang2019stgat,mohamed2020social,Shi2021sgcn,salzmann2020trajectron++,mangalam2020pecnet} in which the first 3.2 seconds (8 frames) are observed and next 4.8 seconds (12 frames) are used for a ground truth trajectory. Additionally, two scenes (Seq-eth, Seq-hotel) of the ETH datasets provide ground-truth group labels. We use them to evaluate how accurately our GP-Graph groups individual pedestrians.

\noindent\textbf{Evaluation protocols.}\quad
For multi-modal human trajectory prediction, we follow a standard evaluation manner, in Social-GAN~\cite{gupta2018social}, generating 20 samples based on predicted probabilistic distributions, and then choosing the best sample to measure the evaluation metrics.
We use same evaluation metrics of previous works~\cite{alahi2016social,gupta2018social,tao2020dynamic,liu2020snce} for future trajectory prediction. Average Displacement Error (ADE) computes the Euclidean distance between a prediction and ground-truth trajectory, while Final Displacement Error (FDE) computes the Euclidean distance between an end-point of prediction and ground-truth. Collision rate (COL) checks the percentage of test cases where the predicted trajectories of different agents run into collisions, and Temporal Correlation Coefficient (TCC) measures the Pearson correlation coefficient of motion patterns between a predicted and ground-truth trajectory. We use both ADE and FDE as accuracy measures, and both COL and TCC as reliability measures in our group-wise prediction. For the COL metric, we average a set of collision ratios over the 20 multi-modal samples.

For grouping measures, we use precision and recall values based on two popular metrics, proposed in prior works~\cite{bisagno2018group,Fernando2018}: A group pair score (PW) measures the ratio between group pairs that disagree on their cluster membership, and all possible pairs in a scene. A Group-MITRE score (GM) is a ratio of the minimum number of links for group members and fake counterparts for pedestrians who are not affiliated with any group.

\newcommand{\ADE}{ADE\raisebox{0.15ex}{\darrow}}
\newcommand{\FDE}{FDE\raisebox{0.15ex}{\darrow}}
\newcommand{\COL}{COL\raisebox{0.15ex}{\darrow}}
\newcommand{\TCC}{TCC\raisebox{0.15ex}{\uarrow}}
\newcommand{\GPR}{\kern1.5ex GR\kern1.5ex}
\newcommand{\GAI}{\kern0.5ex Gain\raisebox{0.15ex}{\uarrow}\kern0.5ex}


\begin{table}[t]
\large
\centering
\resizebox{\linewidth}{!}{
\begin{tabular}{c|c cccccccccc c|c ccccccccc c}
\toprule
\kern4.5em & & \multicolumn{4}{c}{STGCNN} & ~ & \multicolumn{5}{c}{\tbf{GP-Graph\,-\,STGCNN}} & & & \multicolumn{4}{c}{SGCN} & ~ & \multicolumn{5}{c}{\tbf{GP-Graph\,-\,SGCN}} \\ \cmidrule{3-6} \cmidrule{8-12} \cmidrule{15-18} \cmidrule{20-24}
      & & \ADE & \FDE & \COL & \TCC & & \ADE & \FDE & \COL & \TCC & \GAI & & & \ADE & \FDE & \COL & \TCC & & \ADE & \FDE & \COL & \TCC & \GAI \\ \midrule
ETH   & & 0.73 & 1.21 & 1.80 & 0.47 & & \tbf{0.48} & \tbf{0.77} & \tbf{1.15} & \tbf{0.63} & 36.4\% & & & 0.63 & 1.03 & 1.69 & 0.55 & & \tbf{0.43} & \tbf{0.63} & \tbf{1.35} & \tbf{0.65} & 38.8\% \\
HOTEL & & 0.41 & 0.68 & 3.94 & 0.28 & & \tbf{0.24} & \tbf{0.40} & \tbf{2.00} & \tbf{0.32} & 41.2\% & & & 0.32 & 0.55 & 2.52 & 0.29 & & \tbf{0.18} & \tbf{0.30} & \tbf{0.66} & \tbf{0.35} & 45.5\% \\
UNIV  & & 0.49 & 0.91 & 9.69 & 0.63 & & \tbf{0.29} & \tbf{0.47} & \tbf{7.54} & \tbf{0.77} & 48.4\% & & & 0.37 & 0.70 & 6.85 & 0.69 & & \tbf{0.24} & \tbf{0.42} & \tbf{5.52} & \tbf{0.80} & 40.0\% \\
ZARA1 & & 0.33 & 0.52 & 2.54 & 0.71 & & \tbf{0.24} & \tbf{0.40} & \tbf{2.13} & \tbf{0.82} & 23.1\% & & & 0.29 & 0.53 & 0.79 & 0.74 & & \tbf{0.17} & \tbf{0.31} & \tbf{0.62} & \tbf{0.86} & 41.5\% \\
ZARA2 & & 0.30 & 0.48 & 7.15 & 0.39 & & \tbf{0.23} & \tbf{0.40} & \tbf{3.80} & \tbf{0.49} & 16.7\% & & & 0.25 & 0.45 & 2.23 & 0.49 & & \tbf{0.15} & \tbf{0.29} & \tbf{1.44} & \tbf{0.56} & 35.6\% \\ \midrule
AVG   & & 0.45 & 0.76 & 5.02 & 0.50 & & \tbf{0.29} & \tbf{0.49} & \tbf{3.32} & \tbf{0.60} & 35.5\% & & & 0.37 & 0.65 & 2.82 & 0.55 & & \tbf{0.23} & \tbf{0.39} & \tbf{1.92} & \tbf{0.64} & 40.0\% \\ 
SDD   & & 20.8 & 33.2 & 6.79 & 0.47 & & \tbf{10.6} & \tbf{20.5} & \tbf{4.36} & \tbf{0.67} & 38.3\% & & & 25.0 & 41.5 & 4.45 & 0.57 & & \tbf{15.7} & \tbf{32.5} & \tbf{2.59} & \tbf{0.60} & 21.7\% \\
GCS   & & 14.7 & 23.9 & 3.92 & 0.70 & & \tbf{11.5} & \tbf{19.3} & \tbf{1.24} & \tbf{0.73} & 19.2\% & & & 11.2 & 20.7 & 1.45 & 0.78 & & \tbf{7.8}  & \tbf{13.7} & \tbf{0.67} & \tbf{0.79} & 33.8\% \\ \bottomrule
\end{tabular}
}
\resizebox{\linewidth}{!}{
\begin{tabular}{c|c cccccccccc c|c ccccccccc c}
\toprule
\kern4.5em & & \multicolumn{4}{c}{STAR} & ~ & \multicolumn{5}{c}{\tbf{GP-Graph\,-\,STAR}} & & & \multicolumn{4}{c}{PECNet} & ~ & \multicolumn{5}{c}{\tbf{GP-Graph\,-\,PECNet}} \\ \cmidrule{3-6} \cmidrule{8-12} \cmidrule{15-18} \cmidrule{20-24}
      & & \ADE & \FDE & \COL & \TCC & & \ADE & \FDE & \COL & \TCC & \GAI & & & \ADE & \FDE & \COL & \TCC & & \ADE & \FDE & \COL & \TCC & \GAI \\ \midrule
ETH   & & \tbf{0.36} & 0.65 & 1.46 & 0.72 & & 0.37 & \tbf{0.58} & \tbf{0.88} & \tbf{0.77} & 11.0\% & & & 0.64 & 1.13 & 3.08 & 0.58 & & \tbf{0.56} & \tbf{0.82} & \tbf{2.38} & \tbf{0.59} & 27.3\% \\
HOTEL & & 0.17 & 0.36 & 1.51 & \tbf{0.32} & & \tbf{0.16} & \tbf{0.24} & \tbf{1.46} & 0.31 & 32.2\% & & & 0.22 & 0.38 & 5.69 & 0.33 & & \tbf{0.18} & \tbf{0.26} & \tbf{3.45} & \tbf{0.34} & 32.1\% \\
UNIV  & & \tbf{0.31} & 0.62 & 1.95 & 0.69 & & \tbf{0.31} & \tbf{0.57} & \tbf{1.65} & \tbf{0.73} & 7.4\%  & & & 0.35 & 0.57 & 3.80 & 0.75 & & \tbf{0.31} & \tbf{0.46} & \tbf{2.89} & \tbf{0.77} & 19.5\% \\
ZARA1 & & 0.26 & 0.55 & 1.55 & 0.73 & & \tbf{0.24} & \tbf{0.44} & \tbf{1.39} & \tbf{0.82} & 20.3\% & & & 0.25 & 0.45 & 2.99 & 0.80 & & \tbf{0.23} & \tbf{0.40} & \tbf{2.57} & \tbf{0.82} & 11.7\% \\
ZARA2 & & 0.22 & 0.46 & 1.46 & \tbf{0.50} & & \tbf{0.21} & \tbf{0.39} & \tbf{1.27} & 0.46 & 14.3\% & & & 0.18 & 0.31 & 4.91 & 0.55 & & \tbf{0.17} & \tbf{0.27} & \tbf{2.92} & \tbf{0.58} & 13.0\% \\ \midrule
AVG   & & \tbf{0.26} & 0.53 & 1.59 & 0.59 & & \tbf{0.26} & \tbf{0.44} & \tbf{1.33} & \tbf{0.62} & 15.7\% & & & 0.33 & 0.60 & 4.09 & 0.61 & & \tbf{0.29} & \tbf{0.44} & \tbf{2.84} & \tbf{0.62} & 26.4\% \\ 
SDD   & & 14.9 & 28.2 & 0.72 & 0.59 & & \tbf{13.7} & \tbf{25.2} & \tbf{0.35} & \tbf{0.61} & 10.4\% & & & 10.0 & 15.8 & \tbf{0.22} & 0.64 & & \tbf{9.1}  & \tbf{13.8} & 0.23 & \tbf{0.65} & 12.7\% \\
GCS   & & 15.6 & 31.8 & 1.79 & \tbf{0.80} & & \tbf{14.9} & \tbf{30.3} & \tbf{0.81} & \tbf{0.80} & 4.8\% & & & 17.1 & 29.3 & 0.20 & 0.71 & & \tbf{14.2} & \tbf{23.9} & \tbf{0.19} & \tbf{0.72} & 18.4\% \\ \bottomrule
\end{tabular}
}
\vspace{-1mm}
\caption{Comparison between GP-Graph architecture and the vanilla agent-wise interaction graph for four state-of-the-art multi-modal trajectory prediction models, Social-STGCNN\,\cite{mohamed2020social},\,SGCN\,\cite{Shi2021sgcn},\,STAR\,\cite{yu2020spatio} and PECNet\,\cite{mangalam2020pecnet}. The models are evaluated on the ETH\,\cite{5459260}, UCY\,\cite{crowdsbyexample}, SDD\,\cite{robicquet2016learning} and GCS\,\cite{yi2015understanding} datasets. Gain: performance improvement w.r.t FDE over the baseline models, Unit for ADE and FDE: meter, \tbf{Bold}: Best.}
\vspace{-5mm}
\label{tab:gpgraph_trajresult}
\end{table}

\vspace{-3mm}
\subsection{Quantitative Results}
\label{sec:quantitative}
\noindent\textbf{Evaluation on trajectory prediction.}\quad
We first compare our GP-Graph with conventional agent-wise prediction models on the trajectory prediction benchmarks. As reported in~\cref{tab:gpgraph_trajresult}, our GP-Graph achieves consistent performance improvements on all the baseline models. Additionally, our group-aware prediction also reduces the collision rate between agents, and shows analogous motion patterns with its ground truth by capturing the group movement behavior well. The results demonstrate that the trajectory prediction models benefit from the group-awareness cue of our group assignment module.

\noindent\textbf{Evaluation on group estimation.}\quad
We also compare the grouping ability of our GP-Graph with that of state-of-the-art models in~\cref{tab:gpgraph_groupdetect}. Our group assignment module trained in an unsupervised manner achieves superior results in the PW precision in both scenes, but shows relatively low recall values over the baseline models. 

There are various group interaction scenarios in both scenes, and we found that our model sometimes fails to assign pedestrians into one large group when either a person joins the group or the group splits into both sides to avoid a collision. In this situation, while forecasting agent-wise trajectories, it is advantageous to divide the group into sub-groups or singletons, letting them have different behavior patterns. Although false-negative group links sometimes occur during the group estimation because of this, it is not a big issue for trajectory prediction.

To measure the maximum capability of our group estimator, we additionally carry out an experiment with a supervision loss to reduce the false-negative group links. We use a binary cross-entropy loss between the distance matrix and the ground-truth group label. As shown in~\cref{tab:gpgraph_groupdetect}, the performance is comparable to the state-of-the art group estimation models with respect to the PW and GM metrics. This indicates that our learning trajectory grouping network can properly assign groups without needing complex clustering algorithms.

\newcommand{\PW}{PW\raisebox{0.15ex}{\uarrow}}
\newcommand{\GM}{GM\raisebox{0.15ex}{\uarrow}}

\newcommand{\etals}{\,\emph{et\,al}.}

\begin{table}[t]
\large
\centering
\resizebox{\linewidth}{!}{
\begin{tabular}{cc|cccccc|cc}
\toprule
                            &        & Shao\etals\cite{Shao_groupdetection}& Zanotto\etals\cite{Zanotto_groupdetection} & Yamaguchi\etals\cite{yamaguchi2011you} & Ge\etals\cite{Ge_hierarchical_Group} & Solera\etals\cite{Solera_groupdetection} & Fernando\etals\cite{Fernando2018} & GP-Graph & GP-Graph+$\mathcal{S}$ \\ \midrule
\multirow{2}{*}{~Seq-eth}   & ~\PW~ & ~~44.5 / \tbf{87.0}~~ & ~~79.0 / 82.0~~ & ~~72.9 / 78.0~~ & ~~80.7 / 80.7~~ & ~~91.1 / 83.4~~ & ~~\tul{91.3} / 83.5~~ & ~\,\tbf{91.7}\,/ 82.1~~ & ~~91.1 / \tul{84.1}~~ \\ 
                            & ~\GM~ & 69.3 / 68.2 & ~~~-~~~/~~~-~~~ & 60.6 / 76.4 & 87.0 / 84.2 & \tul{91.3} /\,\tbf{94.2} & \tbf{92.5}\,/\,\tbf{94.2} & 86.9 / 86.8 & \tbf{92.5}\,/ \tul{91.3} \\ \midrule
\multirow{2}{*}{~Seq-hotel} & ~\PW~ & 51.5 / 90.4 & 81.0 / 91.0 & 83.7 / 93.9 & 88.9 / 89.3 & 89.1 / 91.9 & 90.2 / \tul{93.1} & \tbf{91.5}\,/ 80.1 & \tul{90.4} /\,\tbf{93.3} \\
                            & ~\GM~ & 67.3 / 64.1 & ~~~-~~~/~~~-~~~ & 84.0 / 51.2 & 89.2 / 90.9 & \tul{97.3} /\,\tbf{97.7} & \tbf{97.5}\,/\,\tbf{97.7} & 84.5 / 80.0 & 96.1 / \tul{96.0} \\
\bottomrule
\end{tabular}
}
\vspace{-1mm}
\caption{Comparison of GP-Graph on SGCN with other state-of-the-art group detection models (Precision/Recall). For fair comparison, the evaluation results are directly referred from~\cite{bisagno2018group,Fernando2018}. $\mathcal{S}$: Use a loss for supervision, \tbf{Bold}: Best, \tul{Underline}: Second best.}
\vspace{-5mm}
\label{tab:gpgraph_groupdetect}
\end{table}

\begin{figure}[t]
\centering
\includegraphics[width=\columnwidth]{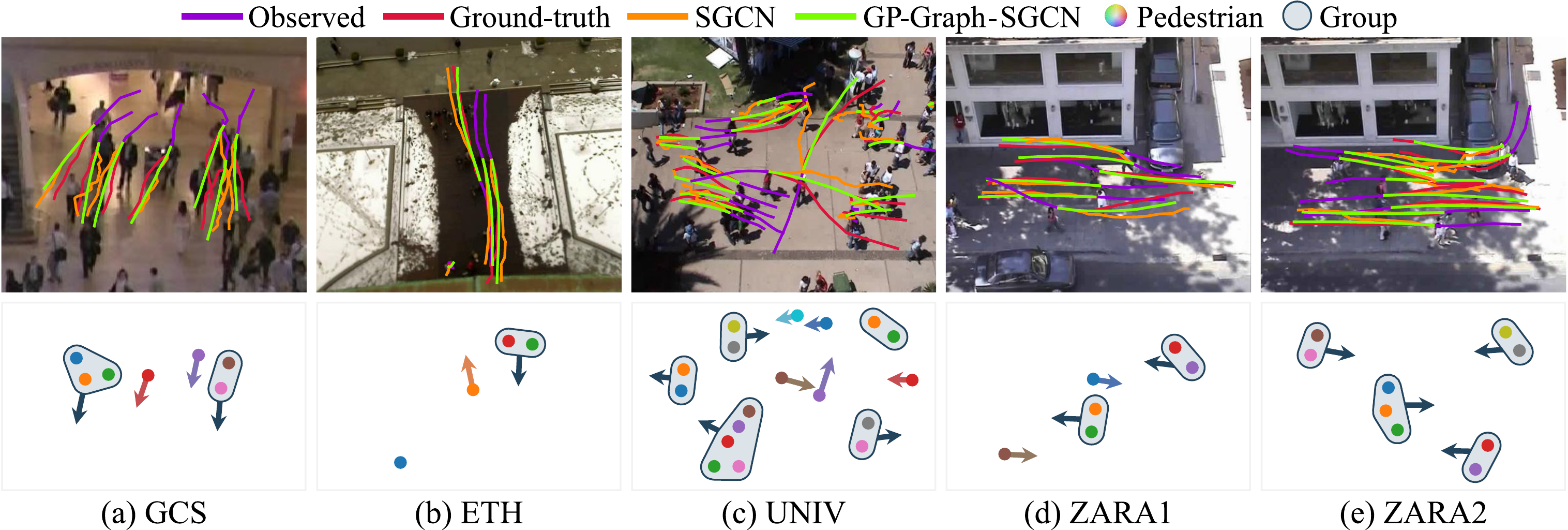}
\vspace{-8mm}
\caption{(Top): Examples of pedestrian trajectory prediction results. (Bottom): Examples of group estimation results on ETH/UCY datasets~\cite{5459260,crowdsbyexample}.}
\vspace{-1mm}
\label{fig:figure5}
\end{figure}

\vspace{-3mm}
\subsection{Qualitative Results}
\label{sec:qualitative}
\noindent\textbf{Trajectory visualization.}\quad
In \cref{fig:figure5}, we visualize some prediction results of GP-Graph and other methods. Since GP-Graph estimates the group-aware representations and captures both intra-/inter-group interactions, the predicted trajectories are closer to socially-acceptable trajectories and forms more stable behaviors between group members than those of the comparison models. \cref{fig:figure5} also shows the pedestrians forming a group with our group assignment module. GP-Graph uses movement patterns and proximity information to properly create a group node for pedestrians who will take the same behaviors and walking directions in the future. This simplifies complex pedestrian graphs and eliminates potential errors associated with the collision avoidance between colleagues.

\begin{figure}[t]
\centering
\includegraphics[width=\columnwidth]{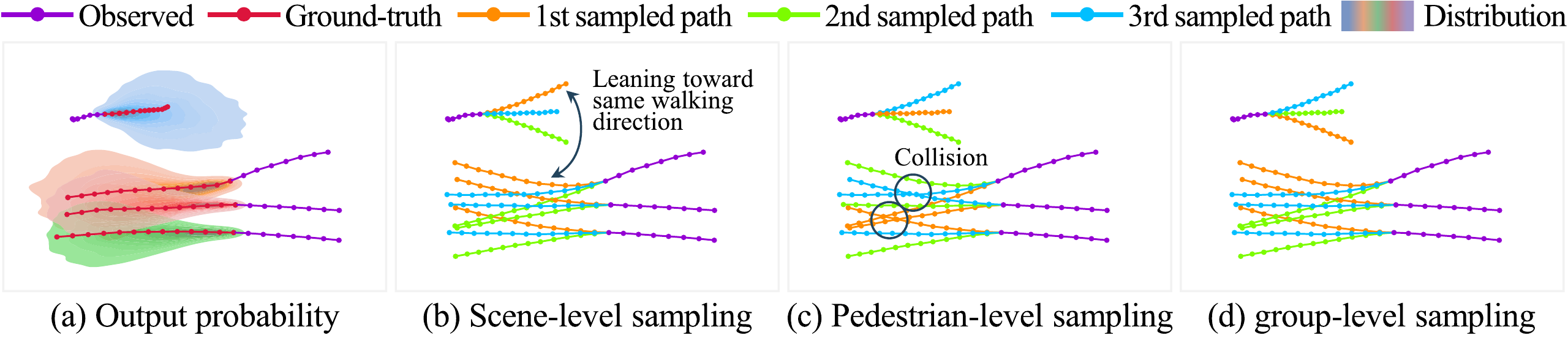}
\vspace{-8mm}
\caption{(a) Visualization of predicted trajectory distribution in ZARA1 scene. (b,c,d) Examples of three sampled trajectories with scene-level, pedestrian-level, and group-level latent vector sampling strategy.}
\vspace{0mm}
\label{fig:figure4}
\end{figure}

\noindent\textbf{Group-level latent vector sampling.}\quad
To demonstrate the effectiveness of the group-level latent vector sampling strategy, we compare ours with two previous strategies: scene-level and pedestrian-level sampling in~\cref{fig:figure4}. Even though the probability maps of pedestrians are well predicted with the estimated group information\,(\cref{fig:figure4}(a)), its limitation still remains. For example, all sampled trajectories in the probability distributions lean toward the same directions\,(\cref{fig:figure4}(b)) or are scattered with different patterns even within group members, which leads to collisions between colleagues\,(\cref{fig:figure4}(c)). Our GP-Graph with the proposed group-level sampling strategy predicts the collaborative walking trajectories of associated group members, which is independent of other groups\,(\cref{fig:figure4}(d)).

\begin{table}[t]
\large
\centering
\resizebox{\linewidth}{!}{
\begin{tabular}{ccccccc}
\toprule
                          & ETH & HOTEL & UNIV & ZARA1 & ZARA2 & AVG \\ \midrule
w/o Pool\&Unpool          & \,\,1.03\,/\,1.69\,/\,0.55\,\, & \,\,0.55\,/\,2.52\,/\,0.29\,\, & \,\,0.70\,/\,\tul{6.85}\,/\,0.69\,\, & \,\,0.53\,/\,1.79\,/\,0.74\,\, & \,\,0.45\,/\,\tul{2.23}\,/\,0.49\,\, & \,\,0.65\,/\,3.02\,/\,0.55\,\, \\
gPool\&gUnpool~\cite{hierarchcal_Gao2019}    & \,0.73\,/\,1.88\,/\,\tul{0.66}\, & \,0.44\,/\,1.78\,/\,\tbf{0.35} & \,\tul{0.44}\,/\,7.67\,/\,\tul{0.78}\, & \,\tul{0.35}\,/\,\tul{1.14}\,/\,\tul{0.84}\, & \,\tul{0.30}\,/\,2.30\,/\,\tul{0.52}\, & \,\tul{0.45}\,/\,\tul{2.96}\,/\,\tul{0.63}\, \\
SAGPool\&gUnpool~\cite{lee2019self}  & \,0.77\,/\,\tbf{1.15}\!\,/\,0.63\, & \,0.40\,/\,2.00\,/\,\tul{0.32}\, & \,0.47\,/\,7.54\,/\,0.77\, & \,0.40\,/\,2.13\,/\,0.82\, & \,0.40\,/\,3.80\,/\,0.49\, & \,0.49\,/\,3.32\,/\,0.60\, \\ \midrule
Group\,Pool\&Unpool       & \,\tul{0.63}\,/\,1.35\,/\,0.65\, & \,\tul{0.30}\,/\,\tul{0.66}\,/\!\,\tbf{0.35} & \tbf{0.42}\!\,/\!\,\tbf{5.52}\!\,/\!\,\tbf{0.80} & \tbf{0.31}\!\,/\!\,\tbf{0.62}\!\,/\!\,\tbf{0.86} & \tbf{0.29}\!\,/\!\,\tbf{1.44}\!\,/\!\,\tbf{0.56} & \tbf{0.39}\!\,/\!\,\tbf{1.92}\!\,/\!\,\tbf{0.64} \\
+Oracle group label       & \tbf{0.62}\!\,/\,\tul{1.27}\,/\!\,\tbf{0.67} & \tbf{0.28}\!\,/\!\,\tbf{0.61}\!\,/\,\tbf{0.35} & -~~~/~~~-~~~/~~~-  & -~~~/~~~-~~~/~~~-  & -~~~/~~~-~~~/~~~-  & -~~~/~~~-~~~/~~~-  \\ \bottomrule
\end{tabular}
    }
\vspace{-1mm}
\caption{Ablation study of various pooling\&unpooling operations on SGCN~\cite{Shi2021sgcn} (FDE/COL/TCC). In the case of our Pedestrian Group Pooling\&Unpooling, we additionally provide experimental results using the ground-truth group labels (Oracle). \tbf{Bold}: Best, \tul{Underline}: Second best.}
\vspace{-5mm}
\label{tab:gpgraph_ablationpool}
\end{table}

\vspace{-2mm}
\subsection{Ablation Study}
\label{sec:ablation}
\noindent\textbf{Pooling\&Unpooling.}\quad
To check the effectiveness of the proposed group pooling\&unpooling layers, we compare it with different pooling methods including gPool~\cite{hierarchcal_Gao2019} and SAGPool~\cite{lee2019self} with respect to FDE, COL and TCC. gPool proposes a top-$k$ pooling by employing a projection vector to compute a rank score for each node. SAGpool is similar to the gPool method, but encodes topology information in a self-attention manner.
As shown in~\cref{tab:gpgraph_ablationpool}, for both gPool and SAGPool, pedestrian features are lost via the pooling operations on unimportant nodes. By contrast, our pooling approach focuses on group representations of the pedestrian graph structure because it is optimized to capture group-related patterns.

\begin{table}[t]
\vspace{2mm}
\large
\centering
\resizebox{\linewidth}{!}{
\begin{tabular}{c|cccccc|cccccc}
\toprule
\multirow{2}{*}{~~\raisebox{-2ex}{\tworow{Varient}{ID}}~~} & \multicolumn{6}{c|}{Components} & \multicolumn{6}{c}{Performance}                                                   \\ \cmidrule(lr){2-7} \cmidrule(lr){8-13}
        & ~~AW~   & ~MB~   & ~GP~   & ~WS~   & ~FG~   & ~GS~~   & ~ETH~         & HOTEL~       & UNIV~        & ZARA1~       & ZARA2~       & AVG~         \\ \midrule
1       & -       & \cmark & -      & -      & -      & -       & ~~0.45 / 0.74~~ & ~0.26 / 0.48~~ & ~0.39 / 0.66~~ & ~0.28 / 0.48~~ & ~0.23 / 0.41~~ & ~0.32 / 0.55~~ \\
2       & -       & -      & \cmark & -      & -      & -       & ~~0.47 / 0.80~~ & \,\tbf{0.17}\,/ \tul{0.31}~~ & ~0.26 / 0.48~~ & ~\tul{0.18} / 0.34~~ & ~\tul{0.16} /\,\tbf{0.29}\,~ & ~0.25 / 0.44~~ \\
3       & -       & \cmark & \cmark & \cmark & -      & -       & ~\,\tbf{0.43}\,/ \tul{0.69}~~ & ~0.20 / 0.37~~ & ~0.25 / 0.47~~ & ~0.19 / 0.35~~ & ~0.17 / \tul{0.32}~~ & ~0.25 / 0.44~~ \\
4       & ~\cmark & \cmark & \cmark & -      & -      & -       & ~~\tul{0.44} / 0.75~~ & ~\tul{0.18} /\,\tbf{0.30}\,~ & \,\tbf{0.23}\,/ \tul{0.43}~~ & ~\tul{0.18} / \tul{0.33}~~ & ~\tul{0.16} / \tbf{0.29}~~ & ~\tul{0.24} / \tul{0.42}~~ \\
5       & ~\cmark & \cmark & \cmark & \cmark & -      & -       & ~\,\tbf{0.43}\,/\,\tbf{0.63}\,~ & ~\tul{0.18} /\,\tbf{0.30}\,~ & ~\tul{0.24} /\,\tbf{0.42}\,~ & \,\tbf{0.17}\,/\,\tbf{0.31}\,~ & \,\tbf{0.15}\,/\,\tbf{0.29}\,~ & \,\tbf{0.23}\,/\,\tbf{0.39}\,~ \\
6       & ~\cmark & \cmark & \cmark & \cmark & \cmark & -       & ~~0.55 / 0.87~~ & ~0.24 / \tul{0.31}~~ & ~0.42 / 0.82~~ & ~0.30 / 0.56~~ & ~0.22 / 0.35~~ & ~0.35 / 0.58~~ \\
7       & ~\cmark & \cmark & \cmark & \cmark & -      & \cmark~ & ~\,\tbf{0.43}\,/\,\tbf{0.63}\,~ & ~\tul{0.18} /\,\tbf{0.30}\,~ & ~\tul{0.24} /\,\tbf{0.42}\,~ & \,\tbf{0.17}\,/\,\tbf{0.31}\,~ & \,\tbf{0.15}\,/\,\tbf{0.29}\,~ & \,\tbf{0.23}\,/\,\tbf{0.39}\,~ \\
\bottomrule
\end{tabular}
}
\vspace{-1mm}
\caption{Ablation study (ADE/FDE). AW, MB, GP, WS, FG and GS respectively denote agent-wise pedestrian graph, intra-group member graph, inter-group group graph, weight sharing among different interaction graph, fixed ratio node reduction of grouping and group-level latent vector sampling respectively. All tests are performed on SGCN. \tbf{Bold}: Best, \tul{Underline}: Second best.}
\vspace{-5mm}
\label{tab:gpgraph_ablationcomponent}
\end{table}

\noindent\textbf{Group hierarchy graph.}\quad
We examine each component of the group hierarchy graph in~\cref{tab:gpgraph_ablationcomponent}. Both intra-/inter-group interaction graphs show a noticeable performance improvement compared to the baseline models, and the inter-group graph with our group pooling operation has the most important role in performance improvement (variants 1 to 4). The best performances can be achieved when all three types of interaction graphs are used with a weight-shared baseline model, which takes full advantage of graph augmentations (variants 4 and 5).

\noindent\textbf{Grouping method.}\quad
We introduce a learnable threshold parameter $\pi$ on the group assignment module in~\cref{eq:pair_distance} because in practice the total number of groups in a scene can change according to the trajectory feature of the input pedestrian node. To highlight the importance of $\pi$, we test a fixed ratio group pooling with a node reduction ratio of 50\%. As expected, the learnable threshold shows lower errors than the fixed ratio of group pooling (variants 5 and 6). This means that it is effective to guarantee the variability of group numbers, since the number can vary even when the same number of pedestrians exists in a scene.

Additionally, we report results for the group-level latent vector sampling strategy (variants 5 and 7). Since the ADE and FDE metrics are based on best-of-many strategies, there is no difference with respect to numerical performance. However, it allows each group to keep their own behavior patterns, and to represent independency between groups, as in~\cref{fig:figure4}.

\vspace{-3mm}
\section{Conclusion}
\vspace{-1mm}
In this paper, we present a GP-Graph architecture for learning group-aware motion representations. 
We model group behaviors in crowded scenes by proposing a group hierarchy graph using novel pedestrian group pooling\&unpooling operations. We use them for our group assignment module and straight-forward group estimation trick. Based on the GP-Graph, we introduce a multi-modal trajectory prediction framework that can attend intra-/inter group interaction features to capture human-human interactions as well as group-group interactions. 
Experiments demonstrate that our method significantly improves performance on challenging pedestrian trajectory prediction datasets.

\noindent\textbf{Acknowledgement} This work is in part supported by the Institute of Information $\&$ communications Technology Planning $\&$ Evaluation (IITP) (No.2019-0-01842, Artificial Intelligence Graduate School Program (GIST), No.2021-0-02068, Artificial Intelligence Innovation Hub), the National Research Foundation of Korea (NRF) (No.2020R1C1C1012635) grant funded by the Korea government (MSIT), Vehicles AI Convergence Research $\&$ Development Program through the National IT Industry Promotion Agency of Korea (NIPA) funded by the Ministry of Science and ICT (No.S1602-20-1001), the GIST-MIT Collaboration grant and AI-based GIST Research Scientist Project funded by the GIST in 2022.

\clearpage
%

\bibliographystyle{splncs04}
\bibliography{egbib}
\end{document}